\documentclass[runningheads]{llncs}
\usepackage[T1]{fontenc}
\usepackage{graphicx}
\usepackage{color}
\newcommand\UrlFont{\color{blue}\rmfamily}

\usepackage{amsmath}
\usepackage{amssymb}
\usepackage{color, colortbl}
\usepackage{array}
\usepackage{pifont}
\usepackage{newfloat}
\usepackage{listings}

\usepackage{subcaption}
\usepackage{graphicx}
\usepackage{makecell}
\usepackage{cuted}
\usepackage{colortbl}
\usepackage{url}
\usepackage{multirow}
\usepackage{booktabs}
\usepackage{mathtools}
\usepackage{hyperref}

\newcommand{\eg}{\textit{e.g.}}
\newcommand{\ie}{\textit{i.e.}}
\newcommand{\xmark}{\ding{55}}%
\newcommand{\cmark}{\ding{51}}%
\begin{document}
%
\title{EPSilon: Efficient Point Sampling for Lightening \\ of Hybrid-based 3D Avatar Generation}
\titlerunning{EPSilon}
%
\author{Seungjun Moon\inst{1} \and
Sangjoon Yu\inst{1} \and
Gyeong-Moon Park\inst{2}\thanks{Corresponding author}}
\authorrunning{S.J. Moon et al.}
%
\institute{KLleon Tech., Seoul, Republic of Korea \and
Kyung Hee University, Yongin, Republic of Korea \\
\email{\{seungjun.moon, sangjoon.yu\}@klleon.io},
\email{gmpark@khu.ac.kr}}
\maketitle              
\begin{abstract}
The rapid advancement of neural radiance fields (NeRF) has paved the way to generate animatable human avatars from a monocular video.
However, the sole usage of NeRF suffers from a lack of details, which results in the emergence of hybrid representation that utilizes SMPL-based mesh together with NeRF representation.
While hybrid-based models show photo-realistic human avatar generation qualities, they suffer from extremely slow inference due to their deformation scheme:
to be aligned with the mesh, hybrid-based models use the deformation based on SMPL skinning weights, which needs high computational costs on each sampled point.
We observe that since most of the sampled points are located in empty space, they do not affect the generation quality but result in inference latency with deformation.
In light of this observation, we propose EPSilon, a hybrid-based 3D avatar generation scheme with novel efficient point sampling strategies that boost both training and inference.
In EPSilon, we propose two methods to omit empty points at rendering; empty ray omission (ERO) and empty interval omission (EIO).
In ERO, we wipe out rays that progress through the empty space.
Then, EIO narrows down the sampling interval on the ray, which wipes out the region not occupied by either clothes or mesh.
The delicate sampling scheme of EPSilon enables not only great computational cost reduction during deformation but also the designation of the important regions to be sampled, which enables a single-stage NeRF structure without hierarchical sampling.
Compared to existing methods, EPSilon maintains the generation quality while using only 3.9\% of sampled points and achieves around $20\times$ faster inference, together with $4\times$ faster training convergence.
We provide video results on \href{https://github.com/seungjun-moon/epsilon}
{\UrlFont{https://github.com/seungjun-moon/epsilon}}.
\end{abstract}
\includegraphics[width=\textwidth]{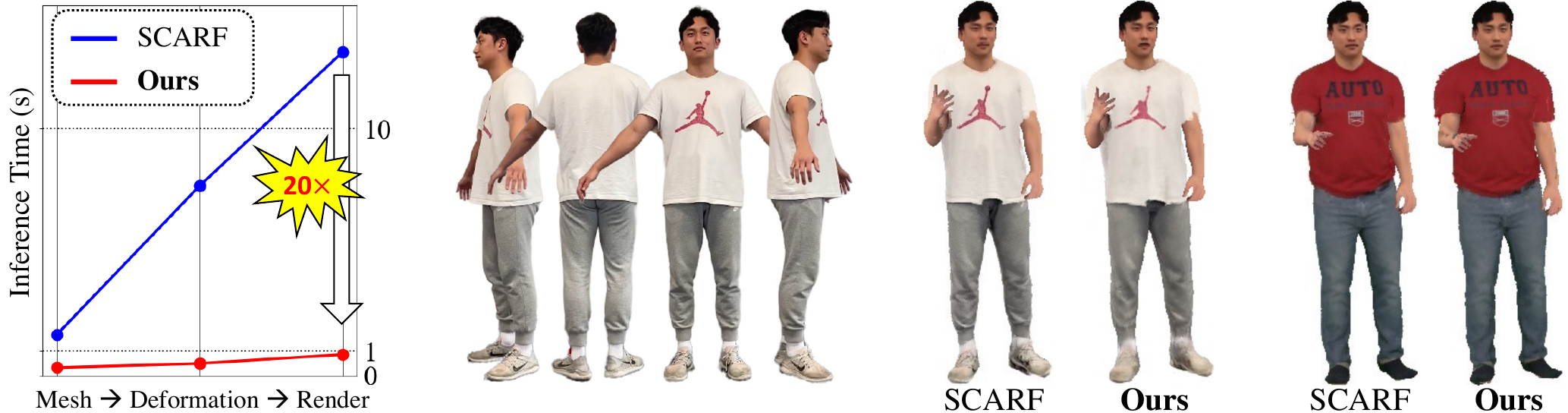}
\captionof{figure}{
\textbf{Comparisons of inference latency and performance with an avatar generation baseline.}
(a) Our method achieves \textbf{20 times} faster inference for generating 300 frames of video, compared to the baseline.
As the baseline does, we only utilize a (b) monocular video and achieve comparable performances on (c) novel pose generation and (d) clothing transfer.
The video can be viewed on our \href{https://github.com/Anonymous4898/epsilon}{project page}.
}
\label{fig:front}

\section{Introduction}
\label{sec:intro}

Along with recent advances in Neural Radiance Fields \cite{mildenhall2021nerf} (NeRF), 3D human avatar generation with deep learning has been prevalent \cite{jiang2022selfrecon,peng2021animatable,chen2021animatable,park2021nerfies,jiang2023instantavatar,feng2022capturing,shen2023x,chen2021snarf,wu2022anifacegan,xu2023omniavatar,wang2022arah,he2021arch++,huang2020arch,xiang2023rendering,wang2022neural,qian2023accelerating,yi2023generalizable}.
The most practical approach is to generate the avatar from a single monocular video \cite{jiang2023instantavatar,xiang2023rendering,feng2022capturing,chen2021animatable,peng2021animatable}, which is cheap and easy to acquire.
However, unless utilizing extra expensive information, \eg, 3D scan \cite{morgenstern2023animatable,shen2023x,su2023caphy}, pure NeRF-based models suffer from a lack of details, \eg, artifacts on fingers and face shapes.
Consequently, hybrid representation \cite{feng2022capturing,feng2023learning} recently emerged, which borrows the detailed human shape representation from the statistically modeled mesh, \eg, SMPL \cite{loper2015smpl,pavlakos2019expressive}, MANO \cite{romero2022embodied}, and FLAME \cite{li2017learning}, while using NeRF to represent the non-human objects, \eg, clothes.
The combination of a strong prior on the human features in a statistical model and the high expressive power of the NeRF model indeed shows exceptional performance in avatar generation \cite{feng2022capturing,feng2023learning}.

While hybrid-based models show remarkably better results than pure NeRF-based models, they suffer from \textit{slow inference time}, \ie, 0.07 FPS \cite{feng2022capturing}.
This is mainly due to the slow \textit{deformation} of hybrid-based models.
Due to the reliance on SMPL mesh, hybrid-based models should deform each sampled point based on SMPL linear blend skinning (LBS) weights, which is relatively slower than MLP-based deformation used in NeRF-based models \cite{chen2023fast,jiang2023instantavatar}.
However, most recent works either focus on boosting the rendering itself \cite{muller2022instant,chen2022tensorf,schwarz2022voxgraf,li2022nerfacc} which does nothing to boost the deformation, or rely on coarse MLP-based deformation \cite{chen2023fast,jiang2023instantavatar} which is incompatible with hybrid representation.
In other words, reducing the deformation cost in hybrid-based models is still under-explored.

In the NeRF rendering scenario, it is well-known that the large portion of the sampled points is \textit{empty}, which does not affect the rendering result due to the low volume density \cite{li2022nerfacc,li2023nerfacc,xu2022point,zhang2023fast}.
Especially for the avatar generation scenario, since the structure of the human body only occupies a small portion of the cubic-shaped rendering space, the portion of empty points is remarkably large \cite{jiang2023instantavatar}.
In light of this observation, we argue that delicate sampling strategies that only select non-empty points can greatly reduce sampling points while preserving the quality.
Since computation costs of deformation are proportional to the number of sampled points, this can effectively boost the inference without performance degradation.
Indeed, in Figure \ref{fig:front}\textcolor{red}{c} and \ref{fig:front}\textcolor{red}{d}, we compare the result which renders with only 3.9\% of the sampled points, \ie, ours, and 100\%, \ie, SCARF \cite{feng2022capturing}, and ours shows the comparable result.
To this end, we argue that efficient point sampling can effectively reduce the computation, even without altering the deformation scheme.

\textbf{Contributions.}
We propose EPSilon, \textbf{E}fficient \textbf{P}oint \textbf{S}ampl\textbf{i}ng for \textbf{l}ightening \textbf{o}f Hybrid-based 3D avatar ge\textbf{n}eration.
EPSilon targets to render only with non-empty points to reduce the computation cost by omitting empty points calculation.
In specific, EPSilon proposes (\lowercase{\romannumeral1}) \textit{empty ray omission} and (\lowercase{\romannumeral2}) \textit{empty interval omission}.
First, using a silhouette of a given SMPL mesh, (\lowercase{\romannumeral1}) classifies the empty ray and omits the rendering process of them.
Second, (\lowercase{\romannumeral2}) narrows the sampling interval to the vicinity of SMPL mesh, to sample points only on the non-empty interval.
Using both sampling strategies, we not only remove hierarchical sampling without notable performance degradation but also reduce the number of sampling rays and points on each ray.
With efficient omitting of the empty points, EPSilon shows comparable results on reconstruction and novel pose generation with the state-of-the-art baseline \cite{feng2022capturing} while achieving 20 times lower latency on video generation, and 4 times faster training speed.
Moreover, EPSilon enables fine pose reenactment and clothing transfer, while fast rendering-based existing baselines \cite{chen2023fast,jiang2023instantavatar} cannot. 
\section{Related Work}

\subsection{Efficient Point Sampling}

Various works focus on efficient point sampling to enhance both rendering quality and speed \cite{mildenhall2021nerf,muller2022instant,jiang2023instantavatar,xu2022point,li2023nerfacc,kurz2022adanerf,yariv2020multiview}.
The key concept of the majority of them is to estimate density \cite{li2023nerfacc} and sample points using it.
However, the estimated density is defined on \textit{canonical space}, while our bottleneck exists on \textit{deformation}, \ie, sending points from observation to canonical space.
Consequently, efficient sampling strategies on canonical space do not reduce the deformation bottleneck.
Rather, we need to sample points efficiently on \textit{observation space}.
InstantAvatar \cite{jiang2023instantavatar} proposes an efficient sampling strategy on observation space, but it requires pre-computation of the density.
In specific, InstantAvatar coarsely samples $64^{3}$ points from observation space and pre-computes their densities to narrow the sampling volume.
Since InstantAvatar uses a fast deformation scheme \cite{chen2023fast}, the cost of the pre-computation is negligible.
However, in hybrid-based models, deformation for $64^{3}$ points are not negligible overheads anymore.
Consequently, we need to narrow down the sampling volume without calculation of deformation in advance.
To the best of our knowledge, EPSilon firstly proposes a method to efficiently sample points in \textit{observation space} within LBS-based models \cite{feng2022capturing,chen2021animatable,feng2023learning}.

\subsection{Deformation of Animatable Avatar Generation}

The majority of NeRF-based avatar generation models \cite{jiang2023instantavatar,feng2022capturing,chen2021animatable,peng2021animatable,alldieck2018detailed,alldieck2018video,chen2021snarf} utilize deformation, which maps the correspondence between the pose-dependent observation space and the pose-independent canonical space in which NeRF is fitted.
Deformation with learnable MLPs \cite{pumarola2021d,park2021nerfies,jiang2023instantavatar} calculates the coordinates mapping with feed-forward calculation, while deformation without learnable parameters \cite{feng2022capturing,feng2023learning,chen2021animatable} utilizes inverses of linear blend skinning (LBS) weights of neighborhood SMPL \cite{loper2015smpl} vertices for the mapping, which is relatively slower than MLP-based deformation.
In the hybrid-based model, since rendered objects should be aligned with SMPL mesh, we need to directly use LBS weights for deformation. \cite{feng2022capturing,feng2023learning}.
Consequently, while achieving outstanding rendered results via hybrid representation, hybrid-based models suffer from slow rendering speed.
With our proposed method, we significantly reduce the latency without performance degradation.

\section{Method}

\begin{figure*}[t]
\begin{center}
\includegraphics[width=\textwidth]{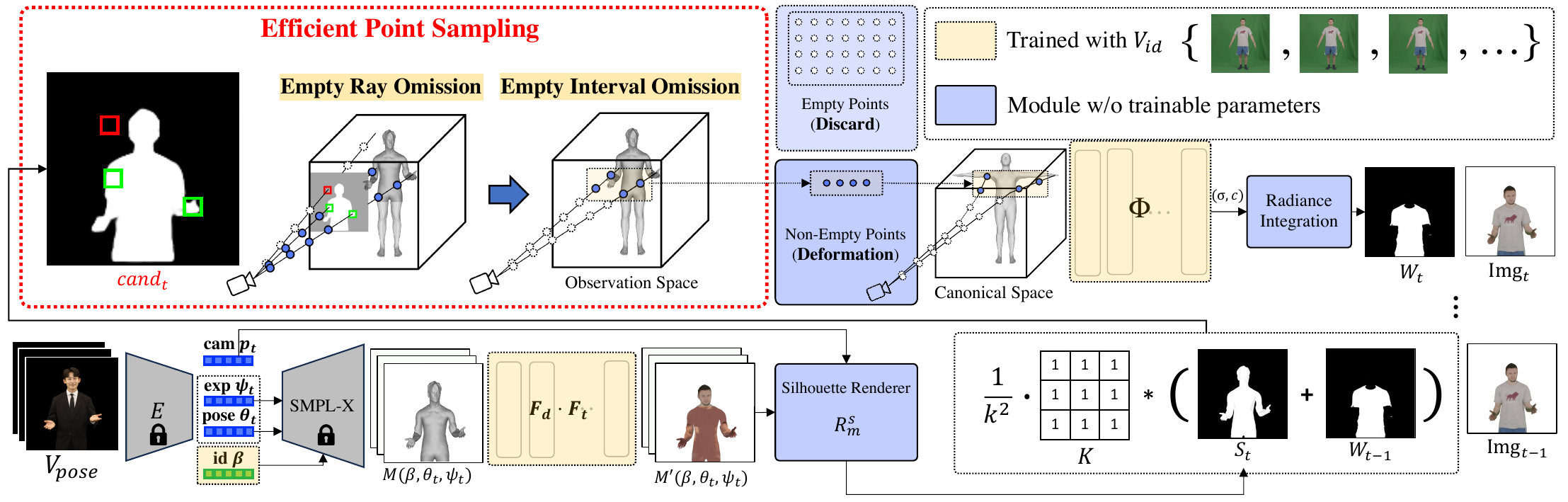}
\end{center}
\caption{
\textbf{Overview of EPSilon.}
From a $t$-th frame in $V_{pose}$, we extract pose, expression, and camera parameter, $\theta_{t}$, $\psi_{t}$, and $p_{t}$, respectively, and combine them with the trained shape parameter $\beta$ to construct the mesh $M(\beta, \theta_{t}, \psi_{t})$.
After passing $M$ through $F_d$ and $F_t$, the mesh $M'(\beta, \theta_{t}, \psi_{t})$ has detailed shape and color.
With $M'$ and $p_{t}$, the model generates the silhouette $S_{t}$, which is added with the dilation of weight map $W_{t-1}$, $W_{t-1}\ast K$, to construct $cand_{t}$.
During the efficient point sampling, empty ray omission filters out empty rays using $cand_{t}$, while empty interval omission forces to sample only in the vicinity of $M'$.
Only with the remaining points, the model progresses deformation to obtain coordinates in the canonical space.
Finally, with the rendering process, we obtain the output frame $\text{Img}_{t}$, and $W_{t}$.
}
\label{fig:architecture}
\vspace{-0.4cm}
\end{figure*}

In this section, we propose novel efficient point sampling strategies for the lightening of hybrid-based 3D avatar generation, named EPSilon.
EPSilon utilizes two new strategies for efficient point sampling: \textit{empty ray omission} and \textit{empty interval omission}.
In Section \ref{sec:method1}, we first introduce the notation and overall structure of EPSilon.
In Section \ref{sec:method2}, we introduce empty ray omission, which prunes redundant rays that only progress through the empty space.
Lastly, in Section \ref{sec:method3}, we introduce empty interval omission, which omits sampling from the interval located far from the mesh.

\subsection{Notation and Architecture}
\label{sec:method1}
We illustrate the total framework of EPSilon in Figure \ref{fig:architecture}.
Our goal is to generate an animatable avatar following the appearance of the human in the given monocular video,  $V_{id}$.
At the training stage, we train the shape parameter $\beta$ of SMPL-X to resemble the human in $V_{id}$, the mesh geometry network $F_{d}: v \rightarrow o$ which calculates the offset of the vertex $v$, the colorization network $F_{t} : v \rightarrow c$ which predicts the RGB color for the vertex $v$, together with NeRF $\Phi$ which implicitly represents the cloth information in $V_{id}$.
Due to the delicate point sampling strategy of EPSilon, we do not need a hierarchical sampling structure in $\Phi$.
We elucidate the detailed training scheme in Appendix.

After training, we generate the animated of the avatar with $V_{pose}$ consisting of $T$ frames.
First, we utilize the pre-trained SMPL-X encoder \cite{zhang2021pymaf,feng2021collaborative} to extract pose, expression, and camera parameters, $\theta_{t}$, $\psi_{t}$, and $p_{t}$, respectively, from frame $t$ of $V_{pose}$ to construct the SMPL-X mesh, $M(\beta, \theta_{t}, \psi_{t})$.
Then, after passing through $F_d$ and $F_t$ successively, the refined mesh $M'(\beta, \theta_{t}, \psi_{t}) = F_t(F_d(M(\beta, \theta_{t}, \psi_{t})))$ is colored and has detailed shapes.
Using the geometry of $M'(\beta, \theta_{t}, \psi_{t})$ in deformation, $\Phi$ renders the clothes that fit the mesh.
Due to the high computation cost of deformation, EPSilon minimizes the number of points to be deformed, with the following strategies: empty ray omission and empty interval omission, which are described in Section \ref{sec:method2} and Section \ref{sec:method3}, respectively.
Using the proposed strategies, we can obtain a valid ray set of frame $t$, $\mathcal{R}_{ERO}(t)$, and a sample set $\mathcal{S}_{(x,y)} = \{r_{m}|r_{m}=o+t_m\cdot d, m\in\{1,\cdots,n_s\}, t_m\in[t_n, t_f]\}$, on the ray $R^{(x,y)}$.
Here, $o$ and $d$ denote the origin and the direction of $R^{(x,y)}$, while $n_s$ denotes the number of sampled points per ray.
The pixel $(x,y)\in \mathcal{R}_{ERO}(t)$ of the rendered image $\text{Img}_{t}^{(x,y)}$ becomes $C(\mathcal{S}_{(x,y)})$, where $C(\mathcal{S}_{(x,y)})$ is a rendering result with $\mathcal{S}_{(x,y)}$.
If $(x,y)\notin \mathcal{R}_{ERO}(t)$, we set $\text{Img}_{t}^{(x,y)}$ to be the constant background color, $c_{back}$.
We use mesh integrated volume rendering \cite{feng2022capturing} to obtain $C(\mathcal{S}_{(x,y)})$:

\begin{align}
  C(\mathcal{S}_{(x,y)}) =& \sum_{i=1}^{n_s-1}\alpha_{i}c_{i}+(1-\sum_{k=1}^{n_s-1}\alpha_k)c, \\
   \text{where } \alpha_i=&\prod_{p=1}^{i-1}exp(-\delta_{p}\sigma_{p})(1-exp(-\sigma_i\delta_i)), \\
  \text{and }
  c=&\begin{cases}
  F_{t}(r_{n_s}^{c,(x,y)}), & \text{if $R^{(x,y)}$ intersects $M'$}\\
  c_{n_s}, & \text{otherwise.}
  \end{cases}
\end{align}

Here, $\delta_{i}=t_{i+1}-t_{i}$, $\Phi(r_i) = (\sigma_i, c_i)$, and $r_{n_s}^{c,(x,y)}$ is canonicalized $r_{n_s}^{(x,y)}$.
Mesh integrated volume rendering sets the last sample on $R^{(x,y)}$, \ie, $r_{n_s}^{(x,y)}$ to be the intersection point when $R^{(x,y)}$ intersects $M'$.
In other words, $t_{n_s}$, \ie, the depth of the $n_s$-th sample of $R^{(x,y)}$, is same with the pixel $(x,y)$ of the depth map $D(M')$, obtained by the differentiable rasterizer \cite{feng2022capturing}.
In the case when $R^{(x,y)}$ does not intersect with $M'$, $D(M')^{(x,y)}$ is set to the constant $t_{far}$.

\subsection{Empty Ray Omission}
\label{sec:method2}
Rays that only pass through the empty space, \ie, \textit{empty rays}, are prevalent because the majority of the volume is empty in the human rendering scenario.
Since the rendering result of empty rays can be substituted to $c_{back}$, calculating deformation and volume integration on empty rays is redundant.
In other words, we can greatly lighten the calculation by predicting empty rays and skipping the rendering on them.
We propose \textit{Empty Ray Omission} (ERO), a novel way to predict empty rays in observation space.
ERO computes a candidate map $cand_{t}\in\mathbb{R}^{H\times W}$ for each frame $t$, which contains a score for each ray $R^{(x,y)}$ in $cand_{t}^{(x,y)}$.
The higher score means the lower probability of the ray being empty.
Using $cand_{t}$, we construct $\mathcal{R}_{ERO}(t)$ as follows:
\begin{equation}
\mathcal{R}_{ERO}(t)=\{(x,y)|cand_{t}^{(x,y)}>\tau_{ERO}\},
\end{equation}
where $\tau_{ERO}$ is a threshold, set to be 0.9 consistently.
We render ${(x,y)}$-th pixel only when $(x,y)\in\mathcal{R}_{ERO}(t)$.
Since $\mathcal{R}_{ERO}(t)$ should contain every pixel that can be non-background, $cand_t$ should delicately assign high scores on pixels which can \textit{possibly} be non-empty.
In other words, pixels in which clothes or SMPL meshes might occupy should be assigned high values in $cand_t$.
We argue that $cand_t$ should be set differently in training and inference, since we assume the continuous motion generation in the inference, while not in the training.

\noindent
\textbf{Training Stage.}
Generally, we can assume that clothes are located near $M'$.
Consequently, we can construct $cand_{t}$ to have high values around the silhouette of $M'$.
In specific, with silhouette renderer $\mathcal{R}^s_m$, we can calculate silhouette of $M'$, $S_{t}=\mathcal{R}^s_m(M', p_{t})\in\mathbb{R}^{H\times W}$, which returns 1 for pixels occupied by $M'$, and 0 for otherwise.
By the convolution between $S_{t}$ and the filter $K_{1}$ with size $k_1\times k_1$ filled with $1/k_{1}^2$, \ie, $S_{t}\ast K_{1}$, pixels near the silhouette become close to 1, \ie, dilation.
Since clothes are located near $M'$, $S_{t}\ast K_{1}$ can be used as $cand_{t}$.
We have to set $k_{1}$ relatively large, \ie, 41, due to the absence of information related to the geometry of clothes.

\noindent
\textbf{Inference Stage.}
Since most animatable avatars target video generation with temporal consistency, we can conjecture the weight map of the previous frame, $W_{t-1}=\sum_{i=1}^{n_s-1} \alpha_i$, is similar to $W_{t}$, which is not available at the training.
In other words, regarding the motion variation of adjacent frames being marginal, we can argue that the dilation of $W_{t-1}$, \ie, convolution between $W_{t-1}$ and the filter $K_{2}$ with size $k_2\times k_2$ filled with $1/k_{2}^2$, contains every pixel that should be rendered at frame $t$.
Moreover, using the prior information of the geometry of clothes from $W_{t-1}$, we can set $k_{2}$ to be smaller than $k_{1}$, \ie, 21, for obtaining more \textit{tight} $cand_{t}$.
We simply add $S_{t}$ and $W_{t-1}$ for more robust candidates construction, and pass convolution with $k_{2}$ for $cand_{t}$.
When $t=1$, we set $cand_{0}^{(x,y)}=S_{0}\ast K_{1}$ since we do not have any pre-computed $W_{t-1}$ to utilize.
See Appendix for detailed hyperparameter settings.

To sum up, we define $cand_{t}$ as below:

\begin{equation}
  cand_{t}=\begin{cases}
    (W_{t-1}+S_{t})\ast K_{2}, & t \in \{2,\cdots,T\}, \\
    S_{t}\ast K_{1}, & t=1 \text{ or training.}
  \end{cases}
\end{equation}

\begin{figure}[t]
\includegraphics[width=\textwidth]{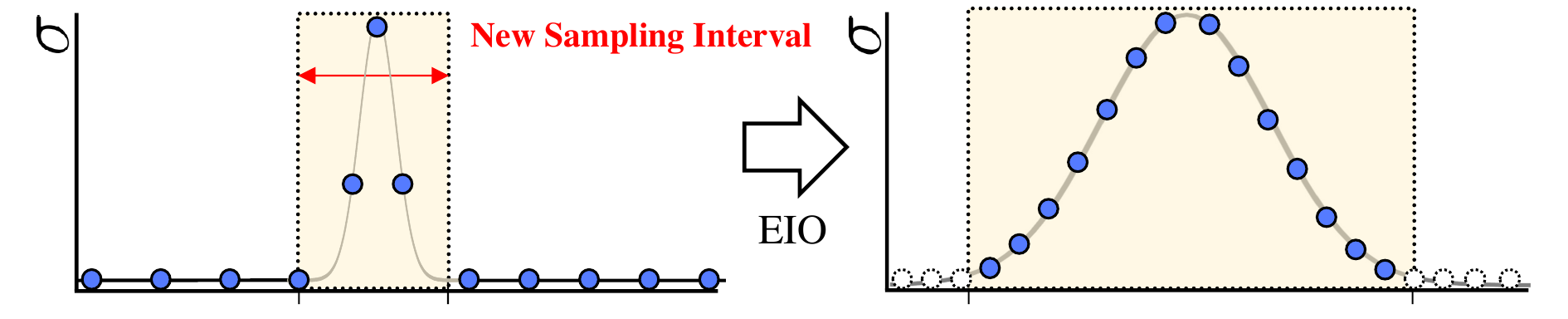}
\caption{
\textbf{Motivation of EIO.}
We empirically find that non-empty points are located in the narrow interval of the ray.
EIO narrows down the sampling interval to discard empty points and render only non-empty points.
}
\label{fig:sampling}
\end{figure}

\subsection{Empty Interval Omission}
\label{sec:method3}

\begin{figure*}[t]
\begin{center}
\begin{subfigure}{0.34\textwidth}
\includegraphics[width=\textwidth]{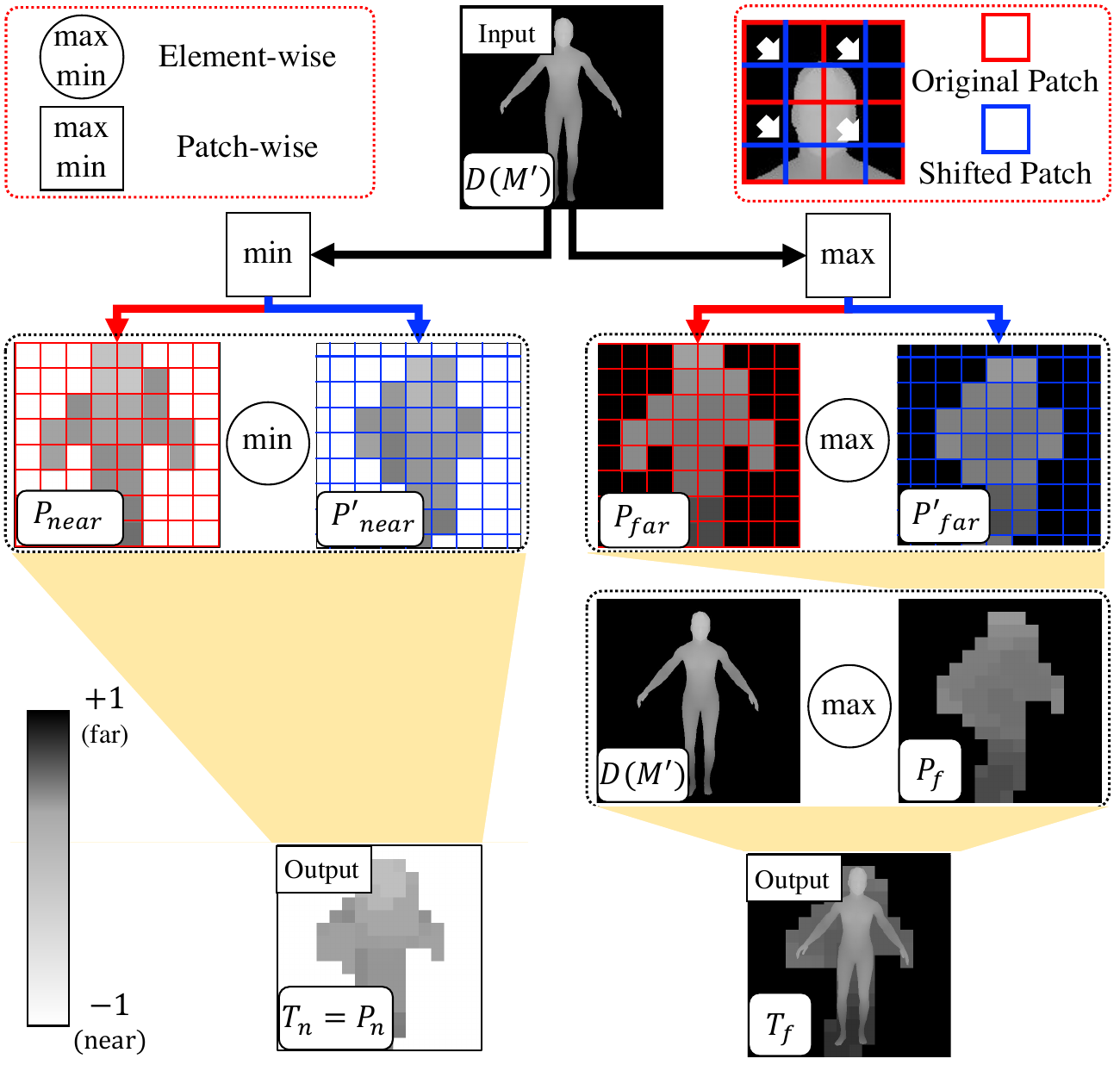}
\caption{Overall scheme.}
\label{fig:eio1}
\end{subfigure}
\begin{subfigure}{0.64\textwidth}
\includegraphics[width=\textwidth]{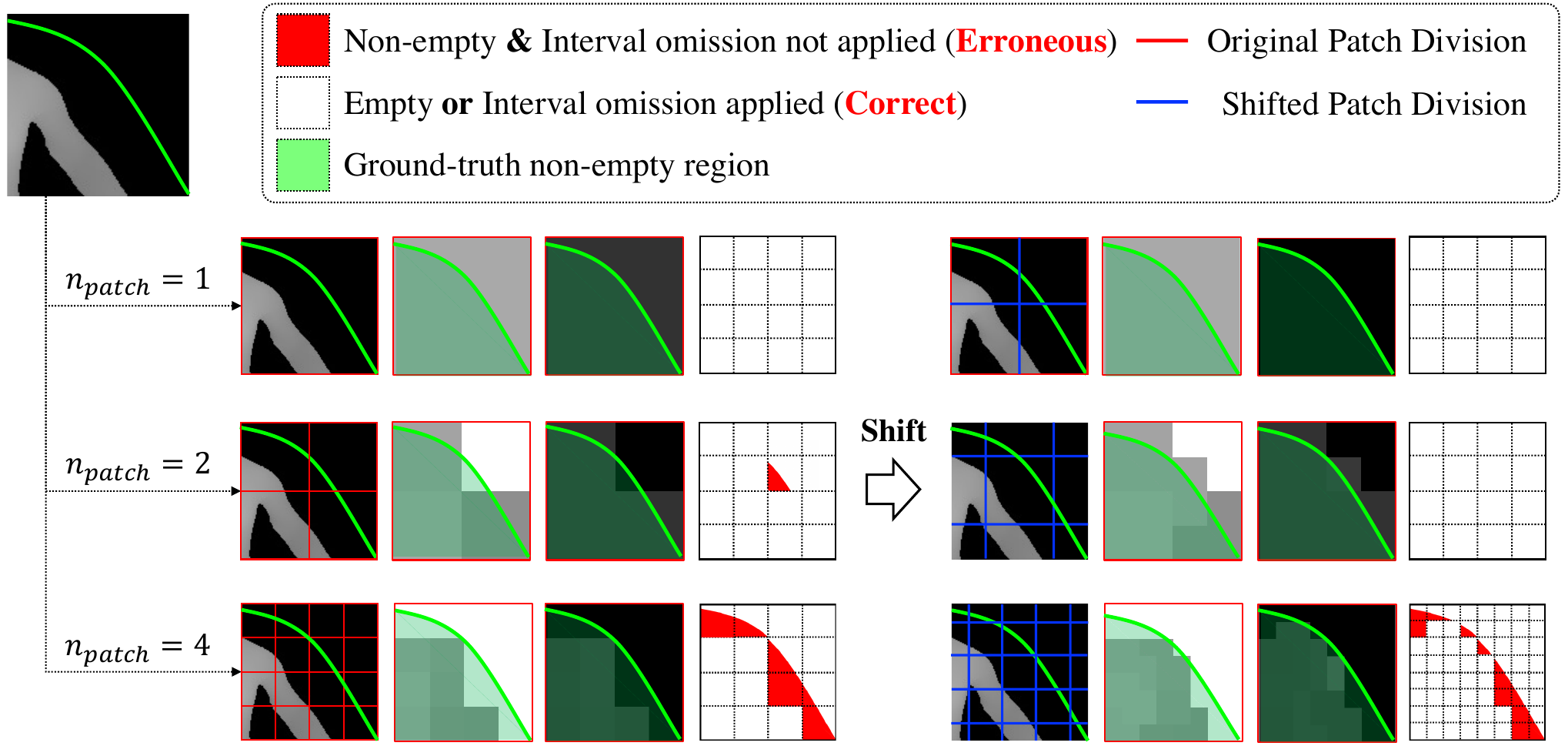}
\caption{Comparison of performance with the patch design.}
\label{fig:eio2}
\end{subfigure}
\caption{
\textbf{Empty interval omission.}
(a) shows the overall scheme of calculating $T_{n}$ and $T_{f}$ in EIO.
Inspired by the shifted window in Swin Transformer, we calculate the efficient sampling interval for pixels in each patch.
(b) shows the effectiveness of window shifting, compared to various size patch designations without window shifting.
In the cases when $n_{patch}=2$ or $4$, \ie, second and third, there exists pixels occupied by clothes (under the green line) but interval omission is not applied, denoted as red.
With decreasing $n_{patch}$, \ie, 1, interval omission is applied to every pixel but very coarsely.
With window shifting, \ie, second column, and the appropriate $n_{patch}$, we can efficiently narrow down the sampling region without missing non-empty pixels from the omission.
}
\label{fig:eio}
\end{center}
\end{figure*}

Though we utilize empty ray omission, numerous points are still empty even on non-empty rays.
We empirically find that the volume density of only a few points is high enough to affect the rendering results, as shown in Figure \ref{fig:sampling}
(See Appendix for the density distribution of real data.).
By delicately narrowing the sampling interval where only non-empty points might be located, we can refrain from the redundant computation of empty points.
Here, we propose a novel method to designate the sampling interval, coined \textit{Empty Interval Omission} (EIO), which targets sample points at only nearby $M'$.
Due to the geometry of clothes, regions far from $M'$ barely have high volume densities \cite{jiang2023instantavatar}.
Using this, EIO narrows down the sampling interval only in the vicinity of $M'$, by using the depth map of $D(M')$.
The original sampling interval of $R^{(x,y)}$ is set to $[t_n, t_f] = [t_{near}, D(M')^{(x,y)}]$, where $t_{near}$ is a constant.
One intuitive approach to narrow the interval is to set $[t_n, t_f] = [D(M')^{(x,y)}-\tau, D(M')^{(x,y)}]$ with a hyper-parametric threshold $\tau$.
This can work when either $S_{t}^{(x,y)}=1$ or $R^{(x,y)}$ is an empty ray, but returns erroneous intervals when $S_{t}^{(x,y)}=0$ and $R^{(x,y)}$ is not an empty ray, \ie, $(x,y)$ is not occupied by $M'$ but occupied by clothes.
In this case, $D(M')^{(x,y)} = t_{far}$ and the sampling interval is fixed to $[t_{far}-\tau, t_{far}]$, \ie, the farthest side of the cubic, without regarding the geometry of clothes.

\begin{table}[t]
\centering
\scriptsize
\begin{tabular}{cc|cc}
\toprule
$n_{patch}$ & Shifted Window & Sampling Volume $\downarrow$ & Error \\
\midrule
- & - & 100.0\% & - \\
\midrule
\multirow{2}{*}{1} & \xmark & 46.7\% & No\\
& \cmark & 66.3\% (+19.6\%) & No \\
\midrule
\multirow{2}{*}{2} & \xmark & 31.8\% & Yes\\
& \cellcolor[gray]{.9} \cmark & \cellcolor[gray]{.9} 34.9\% (+3.1\%) & \cellcolor[gray]{.9} No \\
\midrule
\multirow{2}{*}{4} & \xmark & 17.6\% & Yes \\
& \cmark & 22.4\% (+4.8\%)& Yes \\
\bottomrule
\end{tabular}
\caption{
\textbf{Sampling volumes of various EIO strategies.}
Without the shifted window, setting $n_{patch}=2$ or $4$ occurs in erroneous interval omission though they efficiently reduce the sampling volume.
Combined with the shifted window, only with the slight volume increment, we can remove errors with $n_{patch}=2$.
We set the sampling volume rate to be 100\% when we do not apply EIO.
}

\label{tab:volume}
\end{table}

In light of this observation, EIO utilizes the \textit{patch-based approach} to force the pixels in the vicinity of $M'$ to regard its geometry.
As shown in Figure \ref{fig:eio}, we first divide $D(M')$ into patches with the size $k_{patch}$.
The number of total patches is $n_{patch}^2$, where $n_{patch} = k_{patch}/H$.
Then we calculate the minimum and maximum values of depth within each patch and save the values into $P_{near}$ and $P_{far}$, respectively.
However, when we simply apply the sampling interval of $R^{(x,y)}$ to be $[P_{near}^{(x,y)},P_{far}^{(x,y)}]$, this can yield inaccurate sampling intervals.
In Figure \ref{fig:eio2}, we denote the ground-truth non-empty region, \ie, cloth region, as the green line.
In the patch that does not contain $M'$, the sampling interval becomes $[t_{near}, t_{far}]$ without narrowing.
However, even these patches contain non-empty pixels, and therefore cannot delicately sample points regarding the geometry of $M'$ (red region).
In Figure \ref{fig:eio2}, when $n_{patch}=2$ or $4$, this erroneous interval emerges.
This can be solved by growing up the patch size, \ie, $n_{patch}=1$, but it hinders efficient sampling by containing redundant volumes.
Indeed, as shown in Table \ref{tab:volume}, we can find that the growing $k_{patch}$, \ie, reducing $n_{patch}$, extremely grows the volume for sampling, which decreases the efficiency of interval omission.
To this end, we utilize the \textit{shifted window} in EIO, inspired by Swin Transformer \cite{liu2021swin}.
By shifting windows with $k_{patch}//2$, we can cluster pixels into a single patch, which is separated in pre-computed patch boundaries.
With newly divided patches with shifted windows, we compute $P'_{far}$ and $P'_{near}$.
Final near and far boundaries $\mathcal{P}_{n}$ and $\mathcal{P}_{f}$, respectively, are defined as below:

\begin{equation}
  \mathcal{P}_{n} = \text{min}(P_{near}, P'_{near}), \\
  \mathcal{P}_{f} = \text{max}(P_{far}, P'_{far}).
\end{equation}
As shown in Figure \ref{fig:eio2}, we can remove the erroneous intervals without overly increasing the patch size.
Moreover, as in Table \ref{tab:volume}, we can minimize the volume growth with shifted windows.
However, even with the window shifting, overly increasing $n_{patch}$ still causes erroneous interval omission, as in Figure \ref{fig:eio2}.
Consequently, the appropriate setting of $n_{patch}$ is crucial for the performance.
We elucidate the details for setting $n_{patch}$ in Appendix.

\begin{figure*}[t]
\begin{center}
\includegraphics[width=\textwidth]{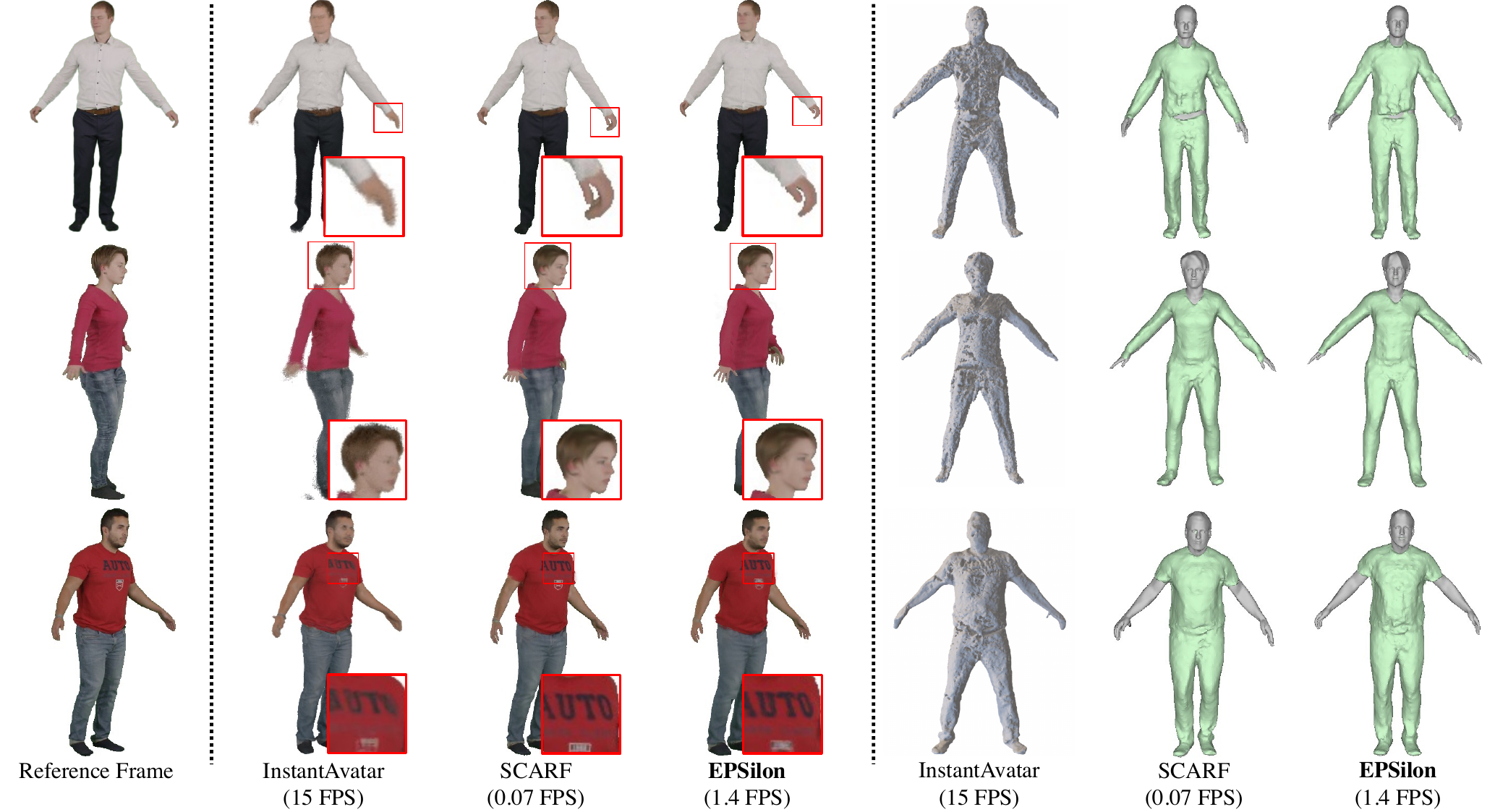}
\end{center}
\caption{
\textbf{Qualitative comparison for reconstruction.}
We compared our model with the pure NeRF-based state-of-the-art baseline which focuses on fast rendering, \ie, InstantAvatar \cite{jiang2023instantavatar}, and hybrid-based baseline with the fine-detailed generation, \ie, SCARF \cite{feng2022capturing}.
While InstantAvatar fails to generate detailed shapes, \eg, blurry clothes and indistinguishable fingers, SCARF and ours successfully reconstruct the details.
We remark that our model rendered \textbf{20 times} faster than SCARF with comparable image and mesh reconstruction qualities.
}
\label{fig:qualitative}
\end{figure*}

Lastly, we combine $D(M')$ with $\mathcal{P}_{f}$ for defining the final far bound $T_{f} = \text{max}(D(M'), \mathcal{P}_{f})$.
We set the final near bound $T_{n} = \mathcal{P}_{n}$.
By narrowing the interval, we can reduce $n_s$ to a quarter of the original without the performance drop.
Despite applying EIO, a few pixels with a large depth difference in $M'$ still need a wide sampling interval.
In this case, we sample with the original $n_s$.
\section{Experiments}
\label{experiments}

In this section, we first briefly introduce the dataset and baselines used for the evaluation.
Then, we compared the quality of generated avatars with baselines in qualitative and quantitative ways.
Especially, with the best-performing baseline, we compared the inference and training time with ours.
Next, we show the contribution of proposed methods, \ie, ERO and EIO, of EPSilon separately, by thorough ablation studies with varying sampling numbers and strategies.
See Appendix for the implementation details of EPSilon and inference environments.

\begin{figure*}[t]
\begin{center}
\includegraphics[width=\textwidth]{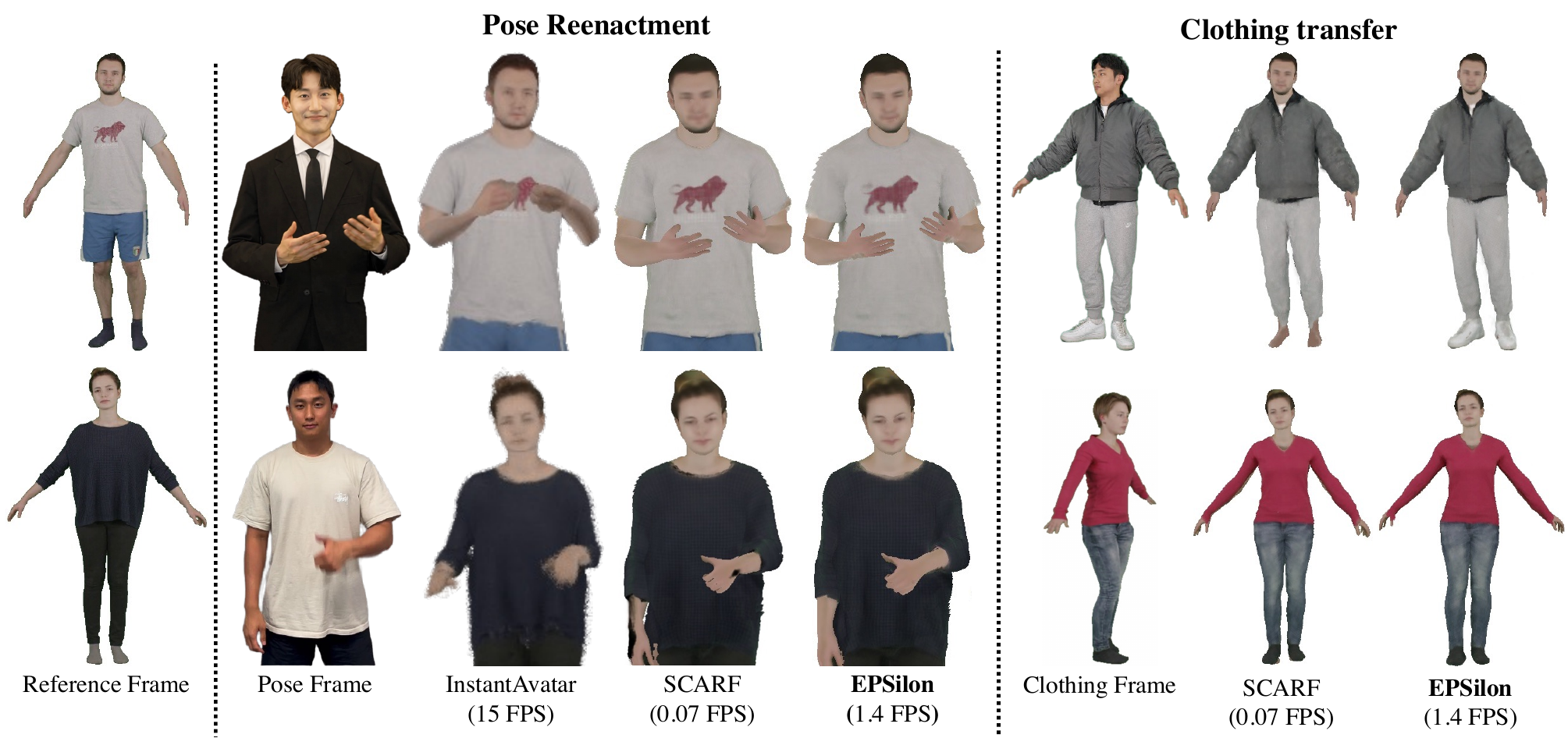}
\end{center}
\caption{
\textbf{Applications of avatar generation.} Our model and baselines enable various applications, \eg, pose reenactment and clothing transfer.
First, pose reenactment enables to generate avatars following a given pose prior.
While our model and SCARF reenacted the fine details of a given pose, InstantAvatar reenacted the coarse pose but not details, \eg, finger shapes.
Moreover, our model and SCARF enable clothing transfer, which can be used on virtual try-on.
With a given clothing prior, both models successfully adapted it to the avatar in the reference frame.
We remark that our model shows \textbf{20$\times$ lower inference latency} than SCARF with comparable application performances.
}
\label{fig:application}
\vspace{-0.2cm}
\end{figure*}
\begin{table*}[ht]
\centering
\scriptsize
\begin{tabular}{cc|cc|cc|cc|cc}
\toprule
\multirow{2}{*}{Models} & \multirow{2}{*}{Rendering} & \multicolumn{2}{c}{male-3-casual} & \multicolumn{2}{c}{male-4-casual} & \multicolumn{2}{c}{female-3-casual} & \multicolumn{2}{c}{female-4-casual} \\
\cline{3-10}
& & PSNR$\uparrow$ & SSIM$\uparrow$ & PSNR$\uparrow$ & SSIM$\uparrow$ & PSNR$\uparrow$ & SSIM$\uparrow$ & PSNR$\uparrow$ & SSIM$\uparrow$ \\
\midrule
\midrule
Anim-NeRF \cite{chen2021animatable} & NeRF & 29.37 & 0.970 & 28.37 & 0.961 & 28.91 & 0.974 & 28.90 & 0.968 \\
InstantAvatar \cite{jiang2023instantavatar} & NeRF & 29.63 & 0.973 & 27.97 & 0.965 & 27.90 & 0.972 & 28.92 & 0.969 \\
\midrule
SCARF \cite{feng2022capturing} & Hybrid &\textbf{30.59} & \textbf{0.977} & \textbf{28.99} & \textbf{0.970} & \textbf{30.14} & \textbf{0.977} & \underline{29.96} & \underline{0.972} \\ 
EPSilon (\textbf{Ours}) & Hybrid & \underline{30.52} & \underline{0.975} & \underline{28.82} & \underline{0.968} & \underline{29.39} & \underline{0.975} & \textbf{30.09} & \textbf{0.974} \\
\bottomrule
\end{tabular}
\caption{\textbf{Quantitative comparison.}
EPSilon consistently showed comparable results with SCARF \cite{feng2022capturing} among various videos while outperforming other SOTA baselines \cite{chen2021animatable,jiang2023instantavatar}.
The bold denotes the best performance, while the underline denotes the second-best performance.
}

\label{tab:quantitative}
\end{table*}
\begin{table}[t]
\centering
\scriptsize
\begin{tabular}{c|cc}
\toprule
& Inference Time (s) $\downarrow$ & Training Time (hr) $\downarrow$ \\
\midrule
SCARF \cite{feng2022capturing} & 13.90 & 6.41 \\
EPSilon (\textbf{Ours}) &  \textbf{0.73} & \textbf{1.55}\\

\bottomrule
\end{tabular}
\caption{
\textbf{Inference and training time.}
The existing model \cite{feng2022capturing} required quite long inference and training time, while EPSilon showed remarkably faster inference \ie, \textbf{$20\times$}, and training, \ie, \textbf{$4\times$} than the baseline model.
}
\label{tab:hybrid}
\end{table}

\begin{table*}[ht]
\scriptsize
\centering
\begin{tabular}{c|c|cc|cc|cccc}
\toprule
Config & \makecell{Hierarchical Sampling} & ERO & EIO & $n_{s,c}$ & $n_{s}$ & \makecell{Sampling Ratio} & PSNR $\uparrow$ & SSIM $\uparrow$ & \makecell{Inference} $\downarrow$ \\
\midrule
A (SCARF) & \checkmark & & & 64 & 96 & 100.0\% & 30.59 & 0.977 & 13.90 \\
B & \checkmark & \checkmark & & 64 & 96 & 53.7\% & 30.58 & 0.977 & 8.78 \\
C & \checkmark & & \checkmark & 64 & 48 & 70.0\% & 30.55 & 0.976 & 10.93 \\
D & \checkmark & \checkmark & \checkmark & 64 & 48 & 46.9\% & 30.54 & 0.974 & 8.18 \\
E & \checkmark & \checkmark & \checkmark & 64 & 24 & 43.4\% & 30.54 & 0.974 & 7.72 \\
\midrule
F & & & & - & 96 & 60.0\% & 30.53 & 0.974 & 7.06 \\
G & & \checkmark & & - & 96 & 13.7\% & 30.53 & 0.974 & 1.73 \\
H & & & \checkmark & - & 48 & 30.0\% & 30.54 & 0.975 & 2.88 \\
I & & \checkmark & \checkmark & - & 48 & 6.9\% & 30.54 & 0.975 & 1.07 \\
\rowcolor[gray]{.9} J (\textbf{EPSilon}) & & \checkmark & \checkmark & - & 28 & 3.9\% & 30.52 & 0.975 & 0.73 \\
\bottomrule
\end{tabular}
\caption{\textbf{Quantitative ablation of EPSilon.}
We compared the PSNR, SSIM, and inference time of variations of EPSilon, on male-3-casual in People Snapshot \cite{alldieck2018video}.
From A to E, we applied ERO and EIO only on the fine NeRF for a fair comparison.
Our proposed method is indeed effective, which showed comparable results only with \textbf{3.9\%} of samples and around \textbf{$20\times$} faster inference.
}
\label{tab:ablation}
\vspace{-0.5cm}
\end{table*}

\subsection{Datasets and Baselines}

We used People Snapshot \cite{alldieck2018video} dataset for both qualitative and quantitative evaluations, along with self-captured data for qualitative evaluation for more various results.
We utilized state-of-the-art avatar generation NeRFs using a monocular video, \eg, Anim-NeRF \cite{chen2021animatable}, InstantAvatar \cite{jiang2023instantavatar}, and SCARF \cite{feng2022capturing}, as the baselines.
Especially, we compared closely with InstantAvatar and SCARF, which show the fastest inference and the best performance among baselines, respectively.

\subsection{Evaluation}
\textbf{Qualitative Evaluation.}
We qualitatively evaluated the generation results in three ways: reconstruction, pose reenactment, and clothing transfer except for InstantAvatar which does not support it.
In Figure \ref{fig:qualitative}, we showed the results for the reconstruction of the given reference frames which are not shown at the training stage.
For a more detailed analysis, we provided not only the image reconstruction results but also the mesh reconstruction results.

Though InstantAvatar \cite{jiang2023instantavatar} generated the coarse clothes or appearances, it failed to generate detailed shapes, \eg, finger shape, and blurs fine details, compared to SCARF or EPSilon.
We conjectured that the missing details of InstantAvatar are related to its erroneous mesh generation.
Indeed, the meshes of InstantAvatar are noisy and suffer from a lack of details, \eg, indistinguishable fingers, which hinders the detailed rendering, while SCARF and EPSilon successfully reconstructed meshes with details.
Moreover, both models enabled separate modeling of body and clothes, which can be used for virtual try-on, as Figure \ref{fig:application}.
Despite the high reconstruction quality of SCARF, its slow inference due to inefficient sampling refrains it from various uses.
On the other hand, EPSilon maintained the reconstruction quality of SCARF and remarkably boosted the inference.

In Figure \ref{fig:application}, we showed results of unseen pose generation.
For a better comparison, we showed the pose reference together with the results.
First, InstantAvatar failed to robustly reenact the pose details.
Since it targets to render frames near real-time, it purely relies on fast NeRF rendering without constructing detailed meshes.
Consequently, it has an inherent limitation in generating fine details, \eg, thumbs up.
On the other hand, SCARF and EPSilon successfully generated fine-detailed poses, while EPSilon showed significantly faster inference.

Finally, we showed the qualitative generation results for clothing transfer.
As mentioned, only hybrid-based models, \eg, SCARF and EPSilon, enable virtual try-on by separately modeling the body and clothes.
In Figure \ref{fig:application}, we showed the generated avatars with changing clothes each other.
EPSilon showed robust cloth transfer results as SCARF, with remarkably faster inference.

\noindent
\textbf{Quantitative Evaluation.}
For the quantitative evaluation, we followed the evaluation setting of baselines \cite{chen2021animatable,feng2022capturing,jiang2023instantavatar}.
We utilized People Snapshot \cite{alldieck2018video} dataset with provided SMPL parameters as initialization.
In Table \ref{tab:quantitative}, we compared the PSNR and SSIM of reconstructed images.
Indeed, hybrid-based models showed consistently better results than pure NeRF-based models on every object.
In specific, EPSilon showed only a 0.7\% drop of PSNR on average and even outperformed SCARF on female-4-casual.
We attributed this to EIO, which delicately narrows the sampling interval where non-empty points are densely located.
In other words, EIO enables a more robust selection of non-empty points even with fewer sampling points.
We compared the training and inference speed of SCARF and EPSilon in Table \ref{tab:hybrid}, which showed that our proposed method is indeed helpful for boosting both training and inference.

\subsection{Ablation Studies}

In this section, we look into the effectiveness of each proposed module as shown in Table \ref{tab:ablation}.
From Config A to E, we only varied the inference scheme, since ERO, EIO, and variation of $n_s$ are applicable in the inference stage.
Also, we fixed the number of sampled points on the ray of the coarse NeRF, \ie, $n_{s,c}$, to the default value of SCARF, \ie, 64.
Similar to Config A-E, we used a single model which is trained without the hierarchical NeRF, from Config F to J.

First, with the reduction of hierarchical sampling (A and E),
we could reduce the sampling ratio by 40\% but still suffered from slow inference, \ie, 7.06 seconds per frame.
By applying ERO, we could reduce the sampling ratio and inference time around $4\times$ (F and G), without performance degradation.
Moreover, we found that applying EIO might slightly increase the performance, even with fewer sampling points (G and I).
We attribute this to the robust interval settings by EIO.
The existing method samples 96 points from the whole cubic space, while EIO forces the model to robustly sample only in the vicinity of the mesh, where non-empty points might be located.
Consequently, we can render with more redundant non-empty points even with fewer sampled points.
In specific objects, \ie, female-4-casual in Table \ref{tab:quantitative}, this even derives the performance gain compared to the baseline.
Moreover, this enables a more aggressive reduction of $n_{s}$ (I and J), \ie, 48$\rightarrow$28, with only a slight performance degradation.
To sum up, EPSilon (J) achieved comparable results with SCARF (A) only with 3.9\% of sampling points and 20 times faster inference.
See Appendix for qualitative comparisons.
\section{Conclusion}

For efficient and elaborate rendering, efficient point sampling on the observation space is essential but under-explored.
We proposed EPSilon, a simple yet effective pipeline for efficient rendering of avatar generation, which consists of empty ray omission and empty interval omission.
We demonstrated that EPSilon maintained the performance of the state-of-the-art model, while the inference latency was reduced to 1/20.
Moreover, our model enabled detailed pose reenactment and clothing transfer, which the models with fast rendering cannot.
We anticipate that our work would be widely applied in research or the industry field, which targets real-time high-fidelity avatar generation.
{
    \small
    \bibliographystyle{splncs04}
    \bibliography{main}
}

\end{document}